\pdfoutput=1
\documentclass[11pt]{article}

\usepackage{ACL2023}

\usepackage{times}
\usepackage{latexsym}

\usepackage[T1]{fontenc}
\usepackage[utf8]{inputenc}

\usepackage{microtype}

\usepackage{inconsolata}

\title{Topic and genre in dialogue}

\author{Amandine Decker$^{1,2}$, Ellen Breitholtz$^2$, Christine Howes$^2$ \and Staffan Larsson$^2$ \\
        $^1$Université de Lorraine \\ \texttt{amandine.decker@loria.fr} \\ $^2$University of Gothenburg \\ \texttt{\{ellen.breitholtz,staffan.larsson\}@ling.gu.se}, \texttt{christine.howes@gu.se}}

\begin{document}
\maketitle
%\begin{abstract}
%In this paper we discuss the role of topics in dialogue and how a hierarchical model of topics could help characterise genres. Such a model would be helpful when it comes to building open-domain dialogue systems that could be fine-tuned for specific tasks.
%\end{abstract}
\section{Introduction}

% Right now approaches that look at certain types of dialogues and define their features but do not necessarily generalise well. Try to take the opposite approach, i.e. we want to find the features that vary in different types of conv so that these features could be defined in a general way and then be adjusted for specific conv. 

% Why do we care about genre and topic?
In this paper we argue that \textit{topic}  plays a fundamental role in conversations, and that the concept is needed in addition to that of genre to define interactions. In particular, the concepts of genre and topic need to be %distinguished and independent %defined separately 
separated and orthogonally defined. This would enable modular,  
%in a way that enables us to for example build
reliable and controllable flexible-domain dialogue systems. %as they determine the subject matter being discussed. 

In communicative activities, genre and topic tend to be interleaved in the sense that the manner in which a particular topic is addressed can differ significantly across various genres. %, and dialogue moves typical of one genre may appear in other genres. 
For instance, a conversation about politics may unfold differently in a formal debate compared to a casual conversation among friends, and a recipe for a dish can be the topic of an instructional dialogue, or a discussion among participants as how to best prepare the dish. 

Analysing the influence of genre on the treatment of topics would allow us to understand how general features of interaction are adapted to specific conversations. In this paper we discuss the treatment of topics and genres in different linguistics theories and how studying the way they influence each other may help designing reliable and controllable open-domain dialogue systems that could be adapted to task-oriented conversations in many different domains.

% What has been done?

% Why are they useful?

%% Why is previous stuff not fine-grained enough?
% What we need to retrieve as topics depends on the task, it can be well defined in task-oriented dialogues but less so in casual conversation. A political debate can seem well cut at first glance but in each argument there is a possibility to diverge. The less task oriented the more complicated + need of a form of hiereachy -> HTM (exists modelisation tools)

% Issue remains that these approaches are quantitative, we do not have explanations on how it works, how the structure works (analysis more than generation, and analysis struggles when it gets less clear cut: topic incoherence + unreasonable structure)

% Conversational patterns

% ``Interaction potential''

\section{Topic %s
and genre %dialogue types
in linguistic theories}

There are several areas of research which aim to categorise interactions in ways that are predictive of their communicative (including linguistic) features. These theories are based on a variety of concepts such as (social) (communicative) \textit{activity} \cite{Allwood2000}, (communicative) \textit{project}, \textit{frame} \cite{LevinMoore1977,Carlson1982}, (language) (dialogue)  \textit{game} \citep{Lewis1979,Ginzburg12}, \textit{genre} \citep{wong-ginzburg-2018-conversational}, etc. 

When defining genres, a frequently used concept is that of \textit{activity} in the context of which language occurs. On Allwood's account an activity type is characterised by the \textit{goals}, \textit{roles}, \textit{artefacts} and \textit{environment} that are associated with it. The carrying out of an activity consists of a number of sub-goals being completed. These may be more or less communicative in nature. For example, instances of the activity type ``Buying/selling coffee in a café'' are made up of sub-goals such as ``conveying which product one wants to order'', ``conveying how much the costumer should pay'', and, finally, ``paying/receiving money''. These sub-goals could be topics in a discussion carrying out the activity type ``Buying/selling coffee in a café'' but they could be organised in many ways with for example all the topics following each other linearly or on the contrary being all embedded in each other. 

Similarly, genres can be seen as a set of actions that must be realised, or a set of questions under discussion that must be resolved \citep{ginzburg2016semantics}, to make a certain interaction successful. In that sense, the genre sets the minimal requirements in terms of outcome for a conversation but it does not say anything on the content of it beyond these requirements. Besides, while genre constrains the surface structure, content plays an important role in the detailed one. 
Formalising topics and understanding the way they can be articulated could help modelling a hierarchical structure of content.

Topics have been discussed in different ways in the literature, the definitions mostly vary by their granularity. A sentence topic \citep{bolinger1952linear, firbas1964defining, halliday1967notes1, givon1983topic} is an element of the sentence, usually a noun phrase, that the sentence comments on \citep{hockett1958course}. \textit{Discourse topics}, on the other hand, are not necessarily explicit. They refer to what a piece of discourse is about, though the formalisation of this ``aboutness'' is debated. Discourse topics have been defined as based on the ``question of immediate concern'' \citep{ochs1983acquiring}, explicitly stated or not, or as ``the proposition or set of propositions that the question of immediate concern presupposes'' \citep{schieffelin1976topic}. Discourse topics are also considered in the frame of certain discourse modelling theories such as Segmented Discourse Representation Theory (SDRT) \citep{Asher2004topic}. Coming up with a theory organising these different levels of granularity would enable us to come up with a hierarchical modelling of topics \citep{whye2006hierarchical}.

\section{Variability in dialogue}

The fulfilment of a conversation goal can be achieved in many different ways, encompassing linguistic and extra-linguistic elements, and their various combinations. Consider a scenario at a café where a customer wishes to order a drink. This goal can be accomplished by pointing at the desired drink, providing a verbal description, employing both actions simultaneously, or in some cases, no action may be necessary if the customer is a regular one with a well-known preference. The diverse range of methods exemplifies the flexibility inherent in achieving conversation goals.

The straightforwardness of attaining conversation goals also varies. Sometimes, intermediate questions need to be resolved before reaching a final decision. For instance, a customer may inquire about the type of milk used in the café and only place their order once they know which drinks are lactose-free. In such cases, the fulfilment of the conversation goal is contingent upon gathering additional information and resolving relevant queries. Such examples show how a straightforward request for action can sometimes turn into something more complex, where information is requested and different alternatives can be discussed and compared.

Conversations may also deviate from a strictly goal-oriented path, allowing for detours and tangential discussions. For instance, while inquiring about a specific product, an individual might share an anecdote related to the product itself. Questions about lavender cookies could trigger memories of holidays in Provence and lead to a spirited debate about the finest variety of lavender or even spark a discussion about the seller's vacation plans. Such diversions from the primary topic rely on the participants' freedom and inclination to explore different avenues within the conversation.

While many conversational goals are associated with a default genre (and related dialogue structures), it sometimes happens that dialogue participants deviate from these defaults. 
The extent to which default structure diverge from the original goals of a conversation is likely influenced by the genre of the conversation and its level of formality or standardisation. 
Additionally, the social aspect of the interaction also plays a central role. It appears that conversations with a greater social orientation tend to afford participants more freedom to deviate from the central goal. 

\section{Application to dialogue systems}

Variability in dialogue is a challenge for general-purpose dialogue models. %The flexibility in dialogue makes it more complicated to propose general-purpose models. %However understanding the relationship between genre, social dynamics, and the flexibility of conversation goals is crucial to uncover the structure of interaction. Moreover, a better understanding of these processes could enable us to model the variations of core features of conversation across different genres which could help improve dialogue systems. 
There may well be an open-ended universe of dialogue genres (language games, dialogue types), which we cannot hope to map out \citep{wittgenstein1953philosophical}. In any case only a limited number of dialogue genres has so far received attention from the dialogue systems / conversational AI community (including industry). 
Having a better understanding of the way topics and genre interact could help creating a more modular and general framework that could be fine-tuned for more specific tasks. 

%In fact building dialogue systems can be regarded as a way of mapping (and organising) the multitude of language games appearing in everyday life into something more tangible. Also, it helps us understand the progress (or lack thereof) of dialogue systems research and development, in terms of which dialogue genres have been covered which have not.

Different dialogue genres will be associated with different kinds of dialogue patterns. %For example, information seeking dialogues will frequently involve question-answer pairs, whereas instructional dialogues will frequently contain requests and acceptances (or rejections) of requests.
In a sense, a notion of dialogue genre is not strictly necessary for dialogue systems. What is needed in each domain is dealing with the dialogue patterns that appear there. However, we believe that the notion of genre can serve as a powerful abstraction, allowing dialogue designers to understand which dialogue patterns are relevant in a domain.%, and which implementation patterns should therefore be used. 

%It is theoretically possible that the variability of dialogue patterns in different domains is so large that the notion of a genre is in the long run not a useful one, and we should be open to this possibility. On the other hand, if the notion of a dialogue genre captures recurring patterns and groups of patterns across domains, dialogue genres may be a useful notion in many ways.

\section{Discussion}

    % Lack of HTM (-> need an annotation guide)

Distinguishing genre and topic and treating the two as orthogonal contributing factors %%   
%    A hierarchical model of topics in interaction would be helpful for several reasons. It
could provide insights regarding the structural analysis of conversation. %and help retrieving relevant information to interpret accurately the context in dialogue. 
In terms of dialogue systems it would improve the adaptation of the model's interventions based on the current topic and its links to the previous ones as well as the type of conversation. However, topic modelling is a complex task even for human annotators \citep{purver2011topic} and creating guidelines to annotate dialogues based on the topics they discuss and the hierarchy between them is not a trivial problem. Such an annotation guide would make it possible to analyse the differences in terms of topical structure between different types of conversations and make dialogue systems adaptable to specific genres of conversations following this analysis.
    % Mapping for different genres can be infinite (thus expensive) task but feel like having a general adaptable framework is less work than building a new framework every time the task changes

\bibliographystyle{acl_natbib}
\bibliography{anthology,custom}

%\appendix

%\section{Example Appendix}
%\label{sec:appendix}

%This is a section in the appendix.

\end{document}